\documentclass{article}

 \usepackage[preprint,nonatbib]{neurips_2020}

\usepackage[utf8]{inputenc} 
\usepackage[T1]{fontenc}    
\usepackage{url}            
\usepackage{booktabs}       
\usepackage{amsfonts}       
\usepackage{nicefrac}       
\usepackage{microtype}      

\usepackage[pdftex]{graphicx}
\usepackage{lipsum}  
\usepackage{dcolumn}

\usepackage{subcaption}  
\usepackage{multirow}    
\usepackage{enumitem}    

\usepackage[boldmath]{numprint}  
\usepackage{siunitx}  
\newcommand\precision{1}
\newcolumntype{L}{>{$}l<{$}} 

\usepackage[table,usenames,dvipsnames]{xcolor} 

\usepackage[export]{adjustbox}  

\usepackage{wrapfig}  
\usepackage{MnSymbol,wasysym}  

\definecolor{control-color}{HTML}{bfbfbf}
\definecolor{showmodelpred-color}{HTML}{5cbca0
}
\definecolor{explanations-color}{HTML}{afacd2}

\definecolor{goodpred}{HTML}{235789}
\definecolor{badpred}{HTML}{a6611a}

\newcommand{\ghuman}{\displaystyle\color{goodpred}\text{Human+}}
\newcommand{\gmodel}{\displaystyle\color{goodpred}\text{Model+}}
\newcommand{\bhuman}{\displaystyle\color{badpred}\text{Human-}}
\newcommand{\bmodel}{\displaystyle\color{badpred}\text{Model-}}
\newcommand{\ghu}{\displaystyle\color{goodpred}\text{H+}}
\newcommand{\gmo}{\displaystyle\color{goodpred}\text{M+}}
\newcommand{\bhu}{\displaystyle\color{badpred}\text{H-}}
\newcommand{\bmo}{\displaystyle\color{badpred}\text{M-}}

\usepackage{etoolbox}
\newtoggle{arxiv}
\toggletrue{arxiv}

\usepackage{xr-hyper}
\usepackage{hyperref}       

\makeatletter
\newcommand*{\addFileDependency}[1]{
  \typeout{(#1)}
  \@addtofilelist{#1}
  \IfFileExists{#1}{}{\typeout{No file #1.}}
}
\makeatother

\newcommand*{\myexternaldocument}[1]{%
    \externaldocument{#1}%
    \addFileDependency{#1.tex}%
    \addFileDependency{#1.aux}%
}

\myexternaldocument{supp}

\title{Are Visual Explanations Useful? \\
A Case Study in Model-in-the-Loop Prediction}

\author{%
    Eric Chu \\
    MIT Media Lab\\
    \texttt{echu@media.mit.edu} \\
    \And
    Deb Roy \\
    MIT Media Lab \\
    \texttt{dkroy@media.mit.edu} \\
    \And
    Jacob Andreas \\
    MIT CSAIL \\
    \texttt{jda@mit.edu} \\
}

\begin{document}

\maketitle

\begin{abstract}
We present a randomized controlled trial for a model-in-the-loop regression task, with the goal of measuring the extent to which (1) good explanations of model predictions increase human accuracy, and (2) faulty explanations decrease human trust in the model. We study explanations based on visual saliency in an image-based age prediction task for which humans and learned models are individually capable but not highly proficient and frequently disagree. Our experimental design separates model quality from explanation quality, and makes it possible to compare treatments involving a variety of explanations of varying levels of quality. We find that presenting model \emph{predictions} improves human accuracy. However, visual \emph{explanations} of various kinds fail to significantly alter human accuracy or trust in the model---regardless of whether explanations characterize an accurate model, an inaccurate one, or are generated randomly and independently of the input image. These findings suggest the need for greater evaluation of explanations in downstream decision making tasks, better design-based tools for presenting explanations to users, and better approaches for generating explanations.
\end{abstract}

\section{Introduction}

While significant research effort has been devoted to automatically explaining decisions from machine learning models,
it remains an open question to what extent these explanations are
useful for humans in downstream applications.
One fundamental assumption underlying much interpretable machine learning research is that more faithful and accurate explanations help people use models more effectively---explanations indicating that models have identified features relevant to the target prediction should increase human confidence in predictions, and explanations indicating that models have latched onto noise or irrelevant features should decrease trust \cite{ribeiro2016should, sundararajan2017axiomatic, lipton2016mythos, doshi2017towards}.
However, most evaluation of explanations has focused on their intrinsic relation to model properties \cite{cai2019effects, narayanan2018humans, ribeiro2016should, sundararajan2017axiomatic}
rather than their effect on human decision-making.
Here we investigate (1) whether explanation quality actually impacts model-in-the-loop human performance, and (2) whether explanation quality impacts human trust in the model. 

\begin{figure}[!t]
    \centering
    \begin{subfigure}[c]{.2\textwidth}
      \includegraphics[width=\textwidth]{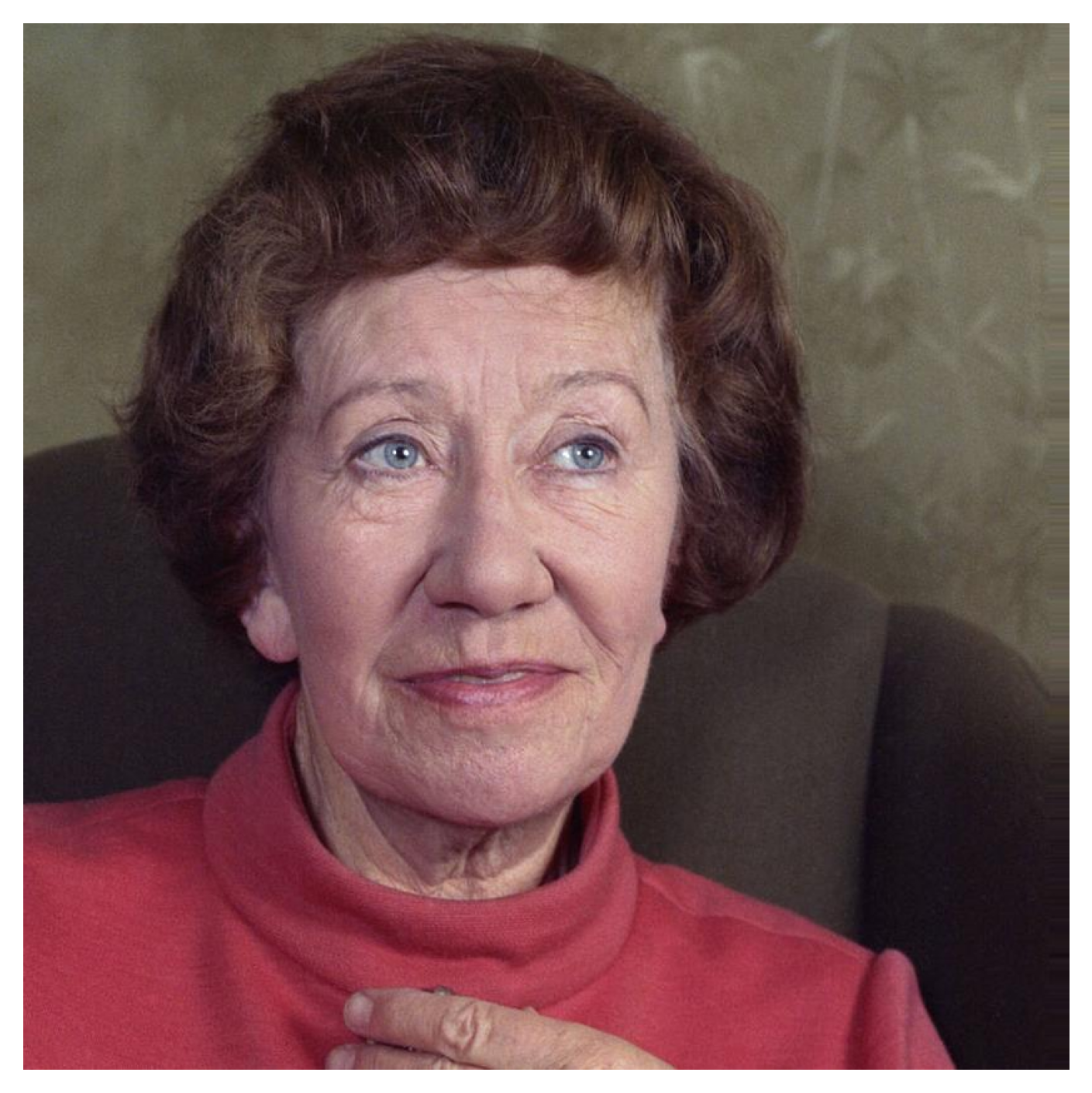}
      \caption*{}
    \end{subfigure}
    \begin{subfigure}[c]{.2\textwidth}
      \includegraphics[width=\textwidth]{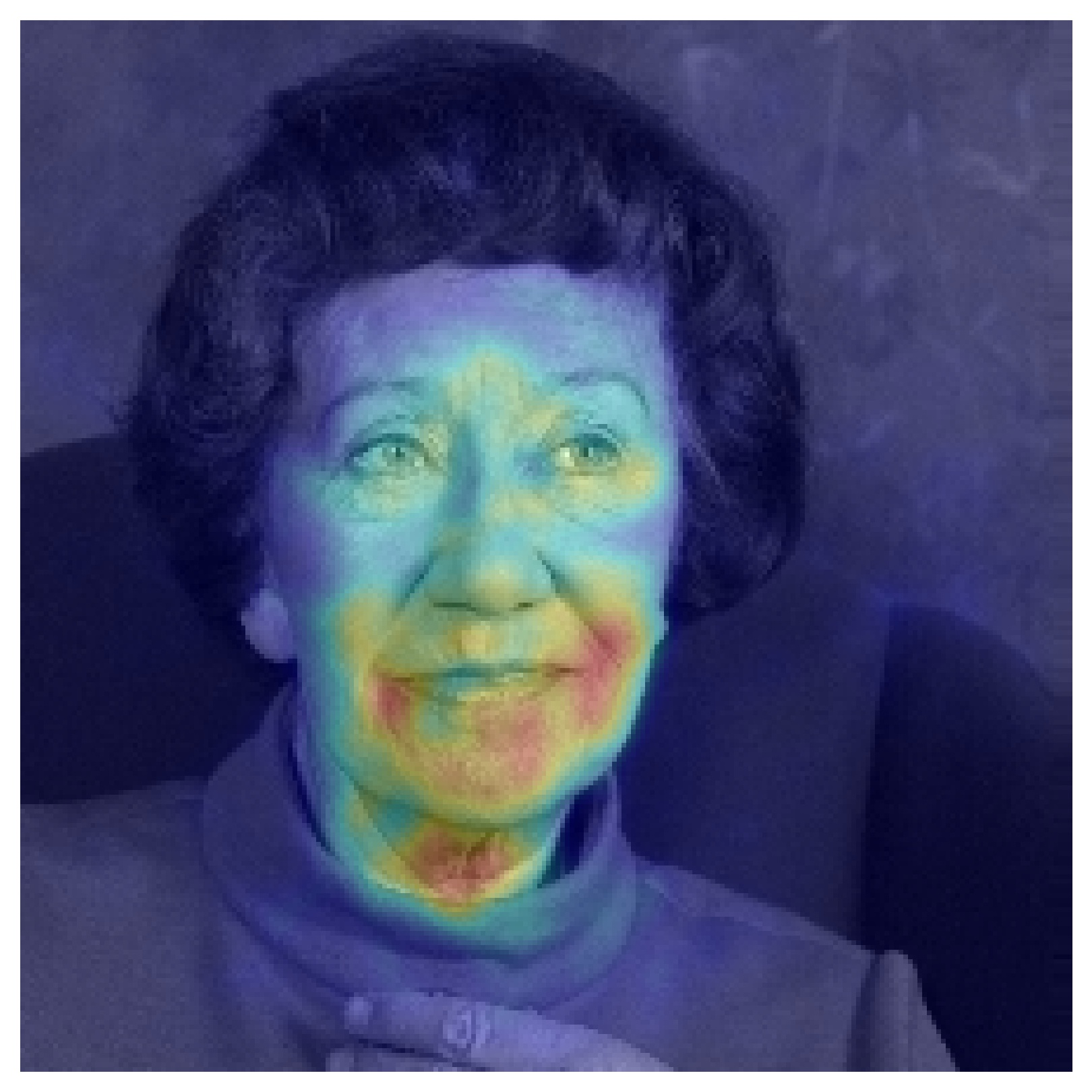}  
      \caption*{}
    \end{subfigure}
    \begin{subfigure}[c]{.2\textwidth}
      \includegraphics[width=\textwidth]{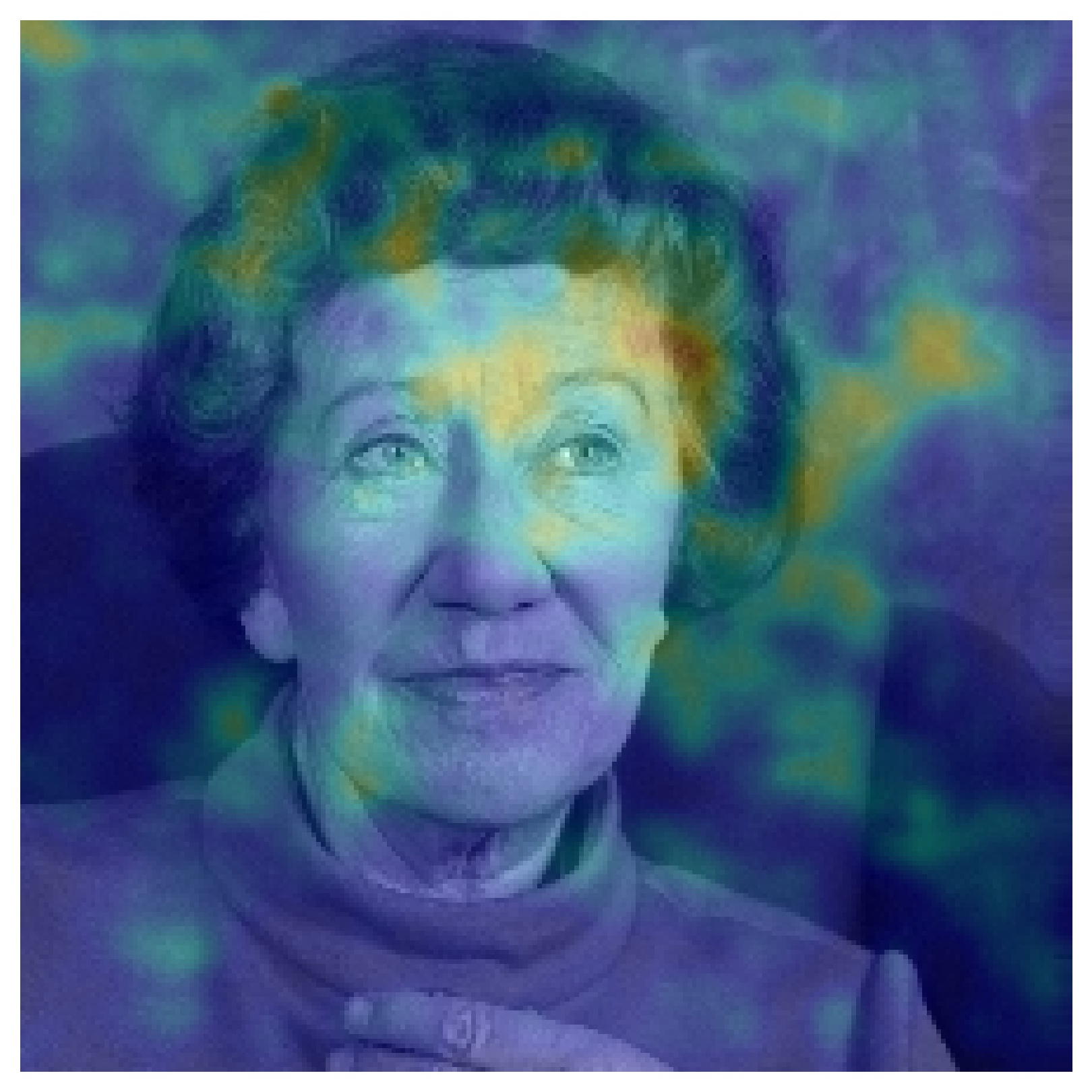}  
      \caption*{}
    \end{subfigure}
    \begin{subfigure}[c]{.2\textwidth}
      \includegraphics[width=\textwidth]{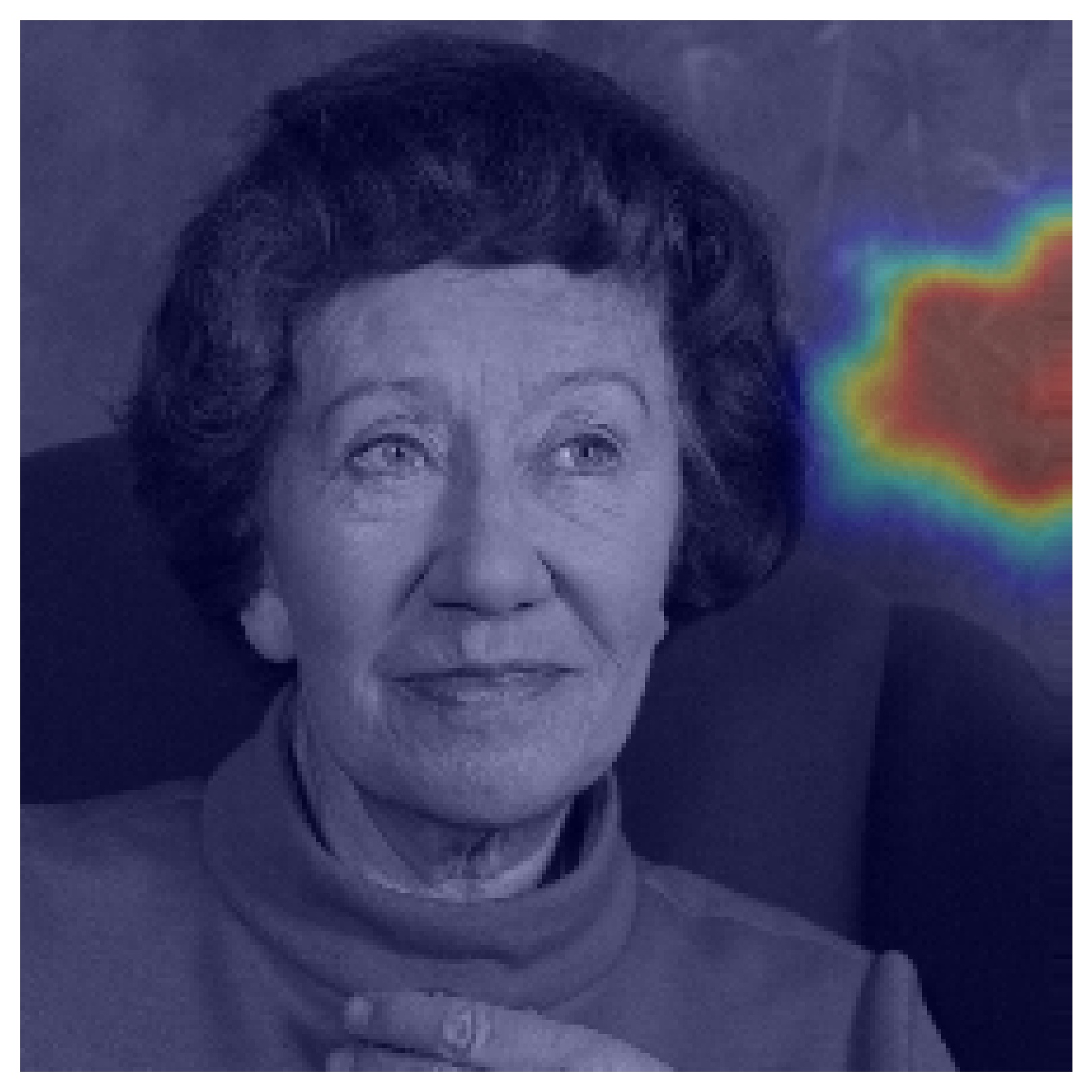}  
      \caption*{}
    \end{subfigure}
    \vspace{-1.5\baselineskip}

    \centering
    \begin{subfigure}[t]{.2\textwidth}
      \includegraphics[width=\textwidth]{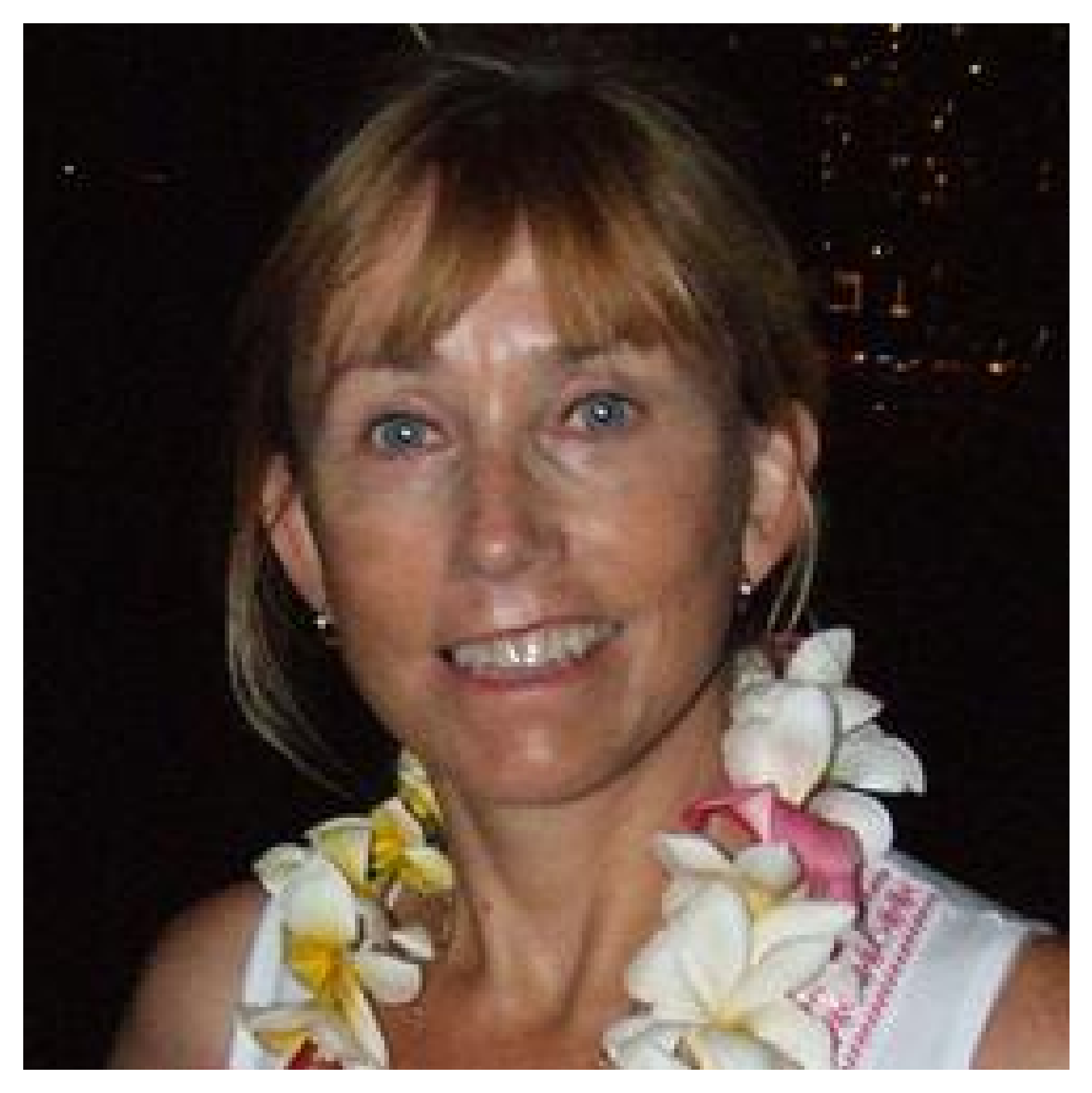}
      \caption{Original image}
    \end{subfigure}
    \begin{subfigure}[t]{.2\textwidth}
      \includegraphics[width=\textwidth]{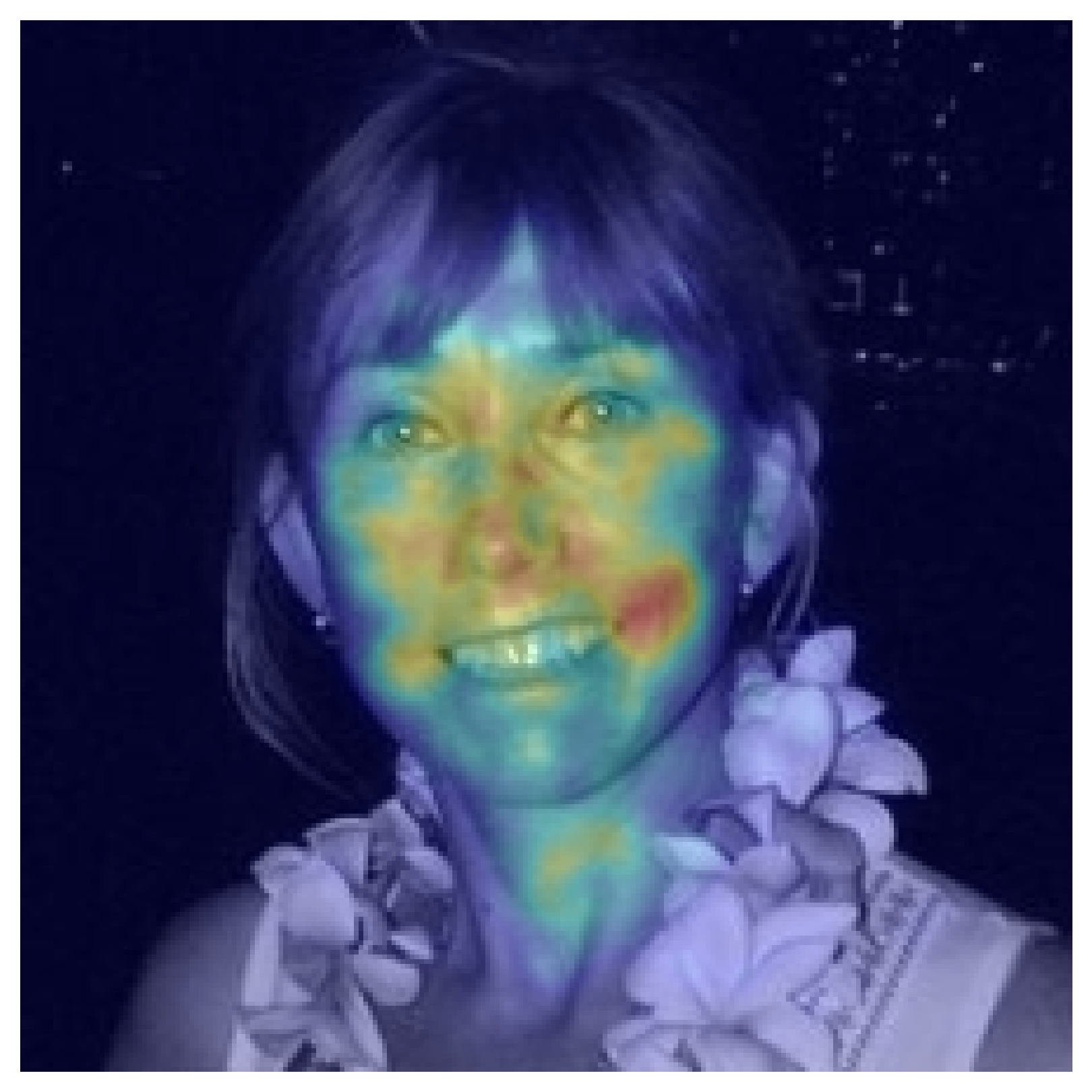}  
      \caption{``Strong''}
    \end{subfigure}
    \begin{subfigure}[t]{.2\textwidth}
      \includegraphics[width=\textwidth]{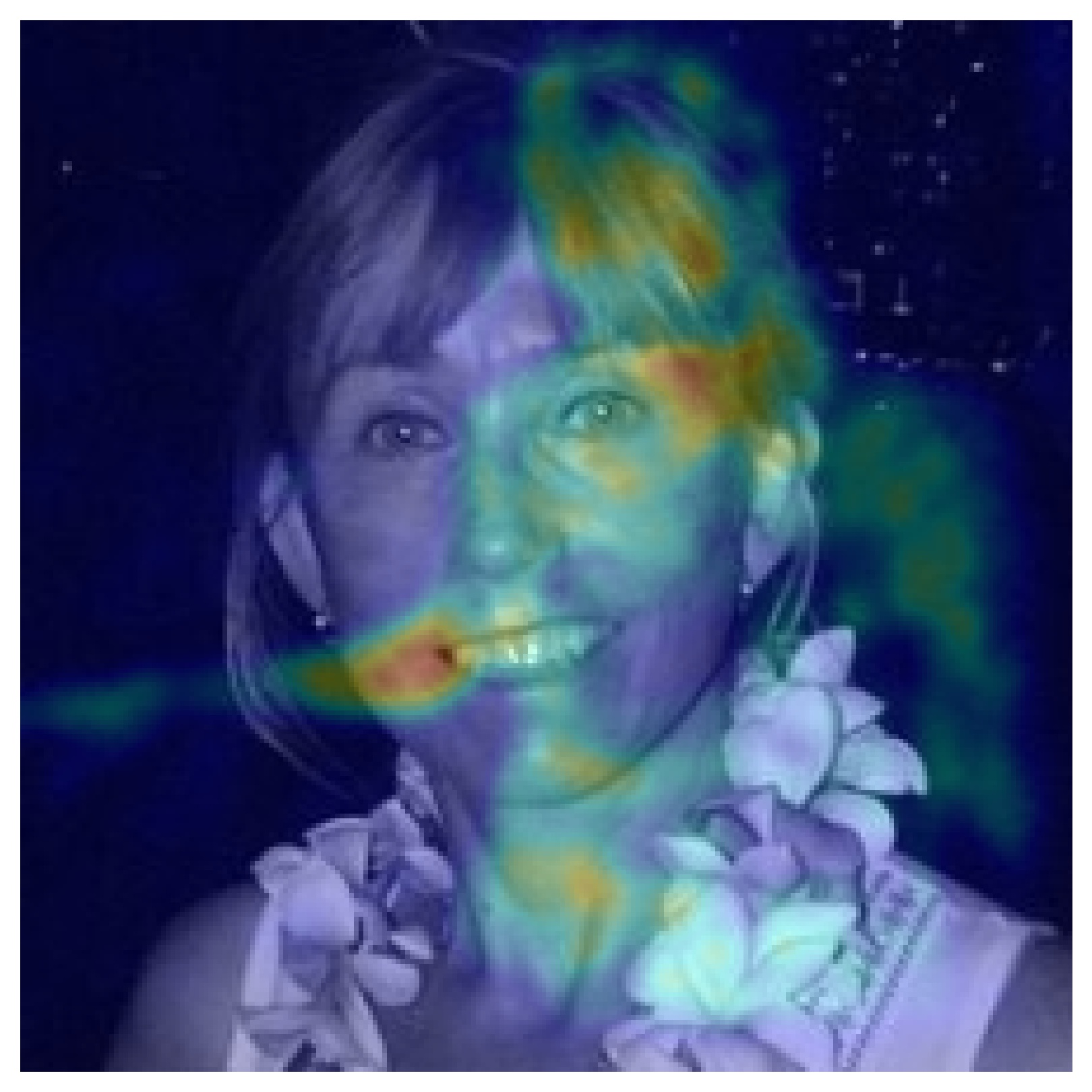}  
      \caption{``Spurious''}
      \label{fig:att-spur}
    \end{subfigure}
    \begin{subfigure}[t]{.2\textwidth}
      \includegraphics[width=\textwidth]{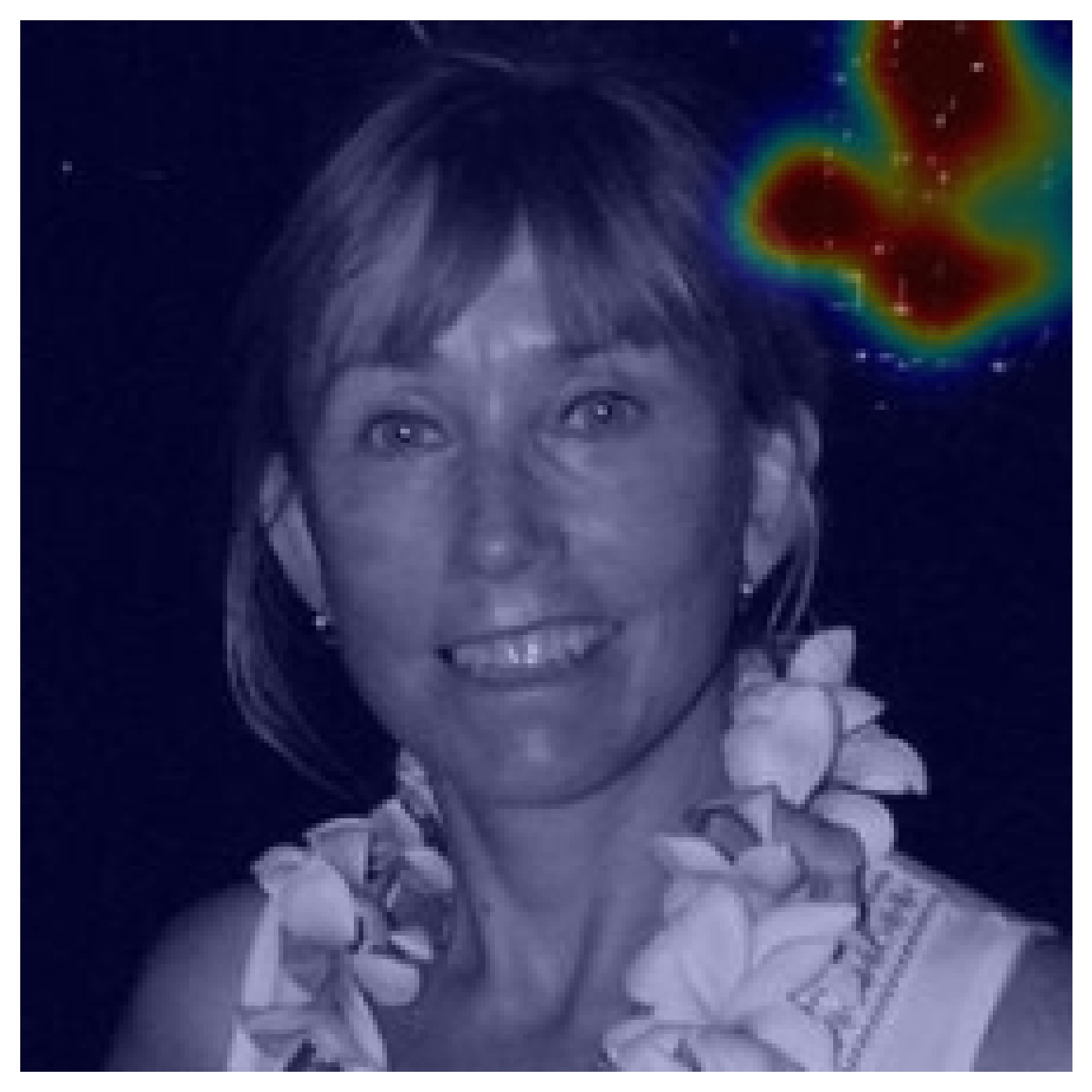}  
      \caption{``Random''}
      \label{fig:att-rand}
    \end{subfigure}
    
\caption{Images and varying saliency maps used in explanation-based treatments in our age prediction task. Participants are shown an image, a prediction from our strong model, and either a ``strong'', ``spurious'', or ``random'' explanation.  The strong explanations are derived from the strong model and focus on details of the face. The spurious explanations are derived from a model trained on data with spurious correlations and tend to focus on both the face and the background. The random explanations are input-agnostic and focus on the background of the image.}.
\label{fig:img-atts}
\end{figure}

We present a randomized controlled trial (RCT) involving model-in-the-loop decision making in a nontrivial perception problem with a modern neural prediction architecture and interpretability method.
Our work follows recent RCTs studying related questions \cite{ binns2018s, feng2019can, green2019principles, lai2020chicago, lai2019human,poursabzi2018manipulating, tan2018investigating, zhang2020effect}, but critically, our setup allows us to isolate the effect of explanations of varying quality in a scenario with more complex models and inputs. 
We include in our experiments both faithful explanations of a high-quality model, via integrated gradients \cite{sundararajan2017axiomatic}, and a variety of ``faulty explanations''---saliency maps from a model trained on data with spurious correlations
and completely uninformative
random explanations, as shown in Figure \ref{fig:img-atts}. We find that neither faulty explanations nor accurate ones significantly affect task accuracy, trust in the model predictions, or understanding of the model. Counter-intuitively, even the obviously unreliable explanations shown in Figure \ref{fig:att-spur} and \ref{fig:att-rand} fail to significantly decrease trust in the model. All of these explanation-based treatments are comparable to each other, and to other treatments such as personifying the model with a name and an image that reveal nothing about its behavior. 
Ultimately, our studies point to the limited effectiveness of pixel-level visual explanations in this model-in-the-loop setting, and motivate research on better explanations, communication of the meaning of explanations, and training of users to interpret them.

\section{Experimental Design}
\label{sec:exp_setup}

\subsection{Task}

We study a model-in-the-loop scenario, where participants are shown an input and a model's prediction, and then asked to make their own guess. Our study examines the age prediction task, where users guess a person's age given an image of their face. We chose this task because (a) both models and humans can perform the task with some proficiency, but with far from perfect accuracy, and (b) it is representative of high-stakes real-world scenarios that use similar models in passport recognition and law enforcement \cite{schuba2020facerecog, passport2020}. A discussion of the ethical concerns of tasks that involve predicting demographic features can be found in the Broader Impact section.

\begin{figure}[!ht]
    \begin{subfigure}[b]{.49\textwidth}
      \centering
      \includegraphics[width=\textwidth, cfbox=lightgray 0.1pt 0.1pt]{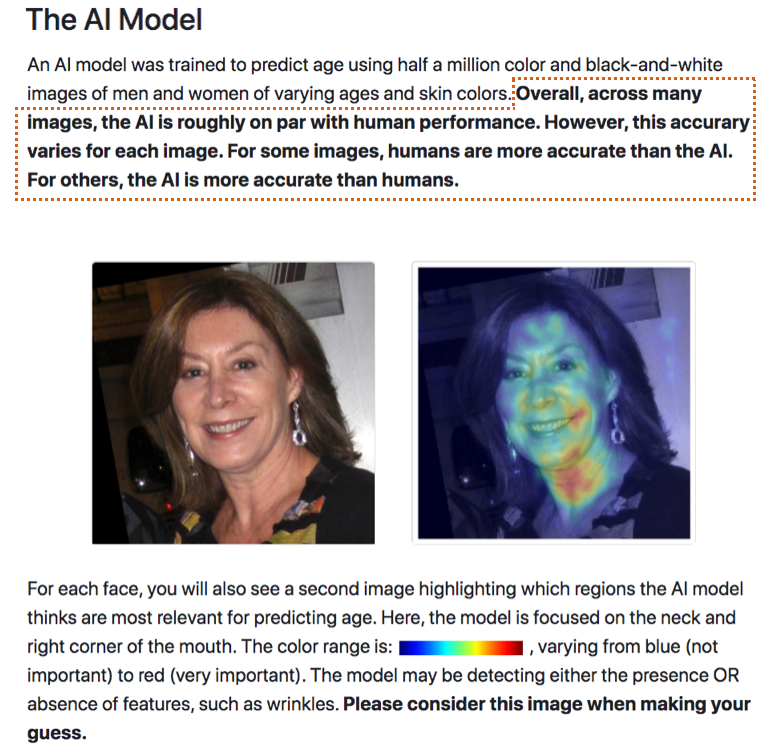}
      \caption{Description of the model and guidelines for interpreting and using the explanations.}
      \label{fig:app-aiguide}
    \end{subfigure}
    \hspace{0.04\textwidth}
    \begin{subfigure}[b]{.5\textwidth}
      \centering
      \includegraphics[width=\textwidth, cfbox=lightgray 0.1pt 0.1pt]{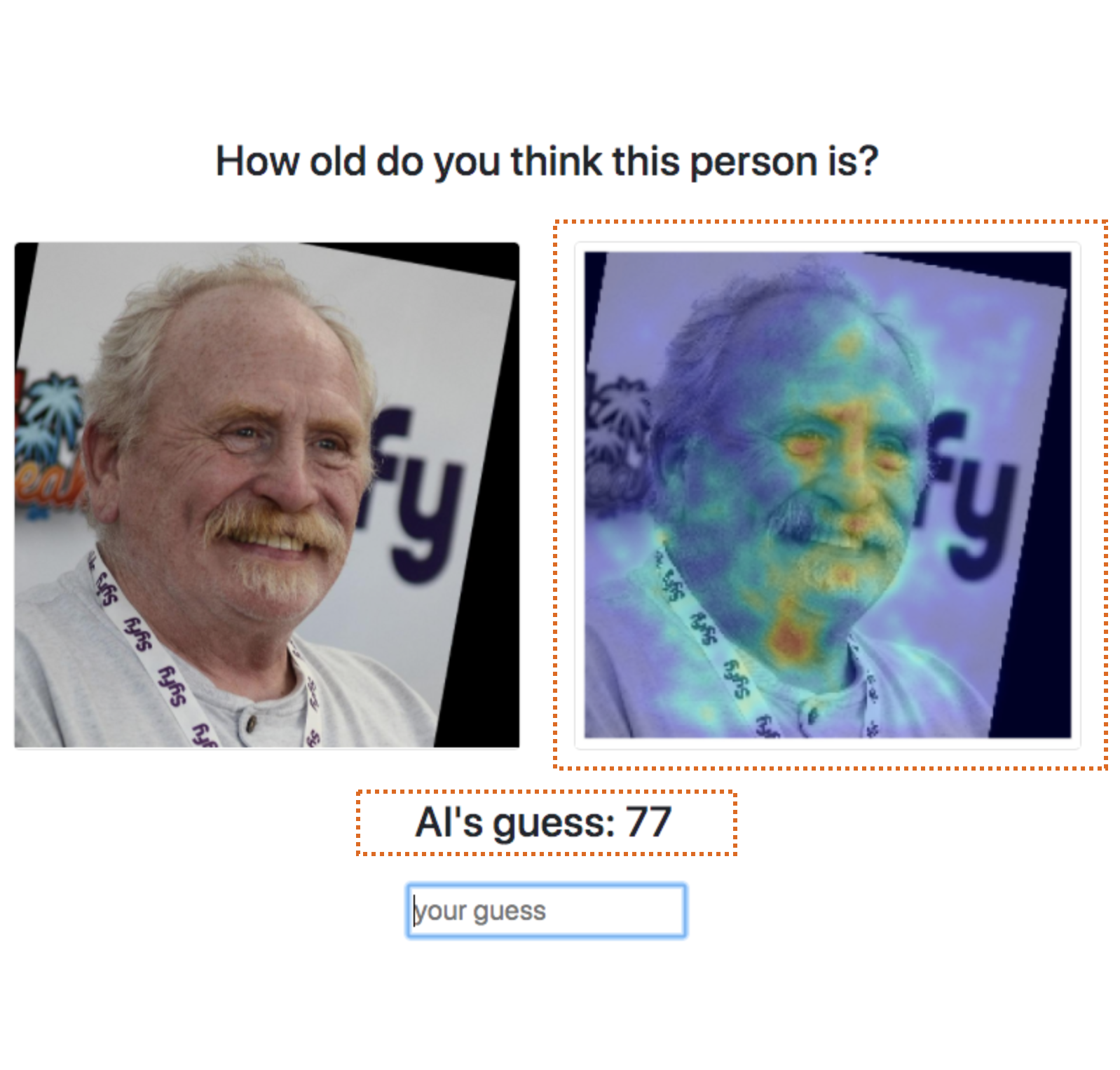}  
      \caption{Users are asked to guess a person's age.}
    \end{subfigure}
    \newline
\caption{Webapp used to run the human study. Elements boxed in orange are shown or not shown depending on the treatment arm. The model prediction is shown in all cases except the control, and in additional treatments discussed in the Appendix. The saliency map in (b) is shown in the Explanation-based treatments. The description of the model performance in (a) is shown in all treatments involving the model, except for two additional treatments discussed in the Appendix.
}
\label{fig:app}
\end{figure}

Users are shown images from the validation and test of the APPA-REAL  dataset \cite{agustsson2017appareal}, which contains 7,591 images with real age labels. We sample images uniformly across ages, and many images are seen by multiple users. We also balance the dataset shown to users by combining equal proportions of images for which the model is more accurate than previously collected human guesses (available in the APPA-REAL data), and vice versa. This allows us to have greater coverage over the more interesting decision space in which the human and model disagree.
The model in our model-in-the-loop task is a Wide Resnet \cite{zagoruyko2016wide} pre-trained on the IMDB-WIKI dataset \cite{rothe2015dex} and fine-tuned on the APPA-REAL dataset. Trained to predict the real ages, this model gets near human-performance with a 5.24 mean absolute error (MAE) on the test set of APPA-REAL, and is more accurate than the previously collected guesses 46.9\% of the time. We call this the ``strong'' model.

\subsection{Treatments}
Users are randomly placed into different experimental conditions, or treatment arms, allowing us to measure the effect of specific interventions. Different treatments vary aspects of the human-AI system, such as whether the model prediction is shown and what kind of explanation accompanies it. Elements of the experiment are shown in Figure \ref{fig:app}, including how model predictions is presented to the participants. The full set of treatments are listed in Table \ref{tab:supp-treatments}. Our main experiments focus on the effect of showing explanations of varying quality alongside model predictions. Additional experiments further exploring showing explanations alone, and the effect of the description of the model performance, are described in the Appendix.

\paragraph{Baseline treatments}
Our two main baselines are (a) the \textit{Control} treatment, in which the user guesses without the help of the model, and (b) the \textit{Prediction} treatment, in which the user guesses with the model prediction shown but without an explanation shown.

\paragraph{Explanation-based treatments}
These treatments show explanations in addition to the model prediction. Our explanations are pixel-wise importances, or saliency maps, calculated using integrated gradients \cite{sundararajan2017axiomatic}. For each pixel, we sum the absolute value of the channel-wise attributions, then normalize these pixel attribution scores by the 98th percentile of scores across that image. Users are given a guide for how to interpret the saliency maps (Figure \ref{fig:app-aiguide}) and also explicitly asked to consider the explanation when making their own guess. 

Explanations of varying quality are shown in Figure \ref{fig:img-atts}. In addition to the strong explanations, we also tested two versions of lower quality saliency maps. Importantly, these varying explanations are all shown with predictions from the same strong model, which isolates the effect of the explanation.

\begin{itemize}[topsep=0pt, itemsep=0ex]
    \item \textit{Explain-strong}. These are saliency maps from the same model whose predictions are shown to users. These explanations tend to focus on, in decreasing order, (a) the areas around the eyes, (b) lines anywhere on the face, and (c) regions around the mouth and nose.
    \item \textit{Explain-spurious}. We train a ``spurious'' model by modifying the APPA-REAL dataset to contain spurious correlations between the background and the label. The area outside the face bounding box is modified by scaling the pixel values by $\alpha$, which is a linear mapping $f(age)$ from the [0,100] age range to a value in [0.25, 5.0]. Saliency maps from the spurious model often focus on both the face and the background. As with all explanation-based treatments, we show the spurious model's saliency map with the predictions from the strong model.
    \item \textit{Explain-random}. We also test completely uninformative, input-agnostic saliency maps that do not focus on the face. To generate these attributions, we first sample an anchor point around the border of the image. We then sample 50 points in a window around the anchor point, which are used as the centers for 2D Gaussians that we then combine. These are similarly normalized and mapped to the same colorscale as the integrated gradients attributions.
\end{itemize}

``Algorithmic aversion'' refers to human loss of trust in a model after seeing it make a mistake \cite{dietvorst2015algorithm, yin2019understanding}. Our question is whether faulty explanations could act as a similar deterrent. A model may be accurate overall but still have undesired behavior. Explanations could expose those deficiencies and serve as a nudge to not trust the model.

\paragraph{Design-based treatments}
We also compare against existing and novel design-based treatments, which vary the description of the model and the way the model's predictions are shown. We are interested in whether these simple framing approaches can be as effective at increasing accuracy or trust as explanation-based treatments. The \textit{Delayed Prediction} treatment tests the effect of anchoring bias \cite{kahneman1982judgment} and was previously shown to work well for improving accuracy in \cite{green2019principles}. We record the initial guess, show the model prediction, then ask for a final guess. The \textit{Empathetic} treatment personifies the model as an AI named Pat, shown in Figure \ref{fig:pat} in the appendix. When a human perceives a computer to be more similar to themselves, they may be more cooperative and find information from the computer to be of higher quality \cite{nass1996can}. We state ``Like every teammate, Pat isn't perfect'' and show Pat next to every prediction. The \textit{Show Top-3 Range} treatment tests a form of uncertainty by showing the range of the models's top 3 predicted ages. The user is told ``The AI's guess will be shown as a range, e.g. 27-29. Sometimes the range may be wider, e.g. 27-35. When the range is wider, this means the AI is more uncertain about its guess.''



\subsection{Metrics} 

We measure and analyze four quantities: (1) the \textbf{error} of the user's guess (the absolute value of the difference between the guess and the \emph{ground-truth} age label), (2) \textbf{trust} (quantified as the absolute value of the difference between the guess and the \emph{model's} prediction), (3) the time
spent making the guess, and (4) answers to post-survey questions on how the model prediction was used in their decision-making process and how reasonable the explanations seemed. Our definition of trust follows previous work operationalizing trust as \textit{agreement} with model predictions \cite{yin2019understanding, zhang2020effect, lai2019human}.



We use a mixed-effects regression model to estimate error and trust as defined above. The model includes fixed-effect terms $\beta_{\textrm{image\_age}}$, the age of the person in the image (which is correlated with the absolute error, $\rho=0.21$, $p < 2.2e^{-16}$), and $\beta_{\textrm{treatment}}$, for each of the treatments. We also include random-effect intercept terms $z_{\textrm{user}}$ and $z_{\textrm{image}}$ to capture effects specific to each image and user. The model is defined as follows, where $\langle$target$\rangle$ is the error or trust defined above.
\begin{equation}
    \label{eq:reg}
    y_{\textrm{$\langle$target$\rangle$}} = \beta_{0} + 
    \beta_{\textrm{treatment}} \cdot x_{\textrm{treatment}} +
    \beta_{\textrm{image\_age}} \cdot x_{\textrm{image\_age}} +
    z_{\textrm{user}} \cdot x_{\textrm{user}} + 
    z_{\textrm{image}} \cdot x_{\textrm{image}} +
    \epsilon
\end{equation}


\subsection{Experiment Details}
We ran experiments on Amazon Mechanical Turk, with 1,058 participants.
Participants were incentivized to perform well on the task by paying top performers a 100\% bonus.
Prior to data collection, we conducted a two-tailed power analysis using the mean and standard deviation of previously collected guesses in the APPA-REAL dataset. In order to detect 1 year differences between treatments, we needed to collect 546 guesses per treatment. We ultimately collected 827.5 guesses (82.75 participants) per treatment, 
which would allow us to detect a 1 year difference of means at the p < 0.05 level with probability 93\%.

\begin{wraptable}{R}{0.52\textwidth}
   \vspace{-1em}
    \caption{Mean absolute error of guesses. Bootstrapped 95\% confidence intervals in parentheses. Results for all treatments are in Appendix Section \ref{sec:supp-results}. We note that the MAE is higher for the ``Model Alone'' case than the MAE stated in Section \ref{sec:exp_setup} because we sample uniformly across the entire age range, and errors are often much larger when the person is older. }
   \label{tab:mae}
      \centering
      \begin{tabular}{lc}
        \toprule
        Treatment Arm & MAE \\
        \midrule
        Control (Human Alone) & 10.0 (9.4 - 10.5) \\
        Model Alone & 8.5 (8.3 - 8.7) \\
        Prediction & 8.4 (7.8 - 9.0) \\
        \midrule
        Explain-strong & 8.0 (7.5 - 8.5) \\
        Explain-spurious & 8.5 (8.0 - 9.1) \\
        Explain-random & 8.7 (8.1 - 9.2) \\
        \midrule
        Delayed Prediction & 8.5 (8.0 - 9.0) \\
        Empathetic & 8.0 (7.6 - 8.5) \\
        Show Top-3 Range & 8.0 (7.4 - 8.5) \\
        \bottomrule
      \end{tabular}
      \vspace{-1em}
\end{wraptable}

\section{Analysis and Results}
The overall mean absolute errors per treatment are shown in Table \ref{tab:mae}. For more detailed analysis, we control for image- and user-specific effects using the regression model in \autoref{eq:reg}, shown in Figures \ref{fig:reg-err} and \ref{fig:reg-trust}.
This allows us to more precisely quantify the additional effect of explanation-based treatments on error relative to a human-only control (\autoref{fig:reg-err}) and the effect of explanations on trust relative to a prediction-only control (\autoref{fig:reg-trust}). 
In addition to a regression on the overall data, we also analyzed subsets of the data, defined by the mean of all human and model errors (8.65). For example, the ``good human, bad model'' subset, denoted $\ghuman\bmodel$ and $\ghu\bmo$ for short, is the set of datapoints where the human guess was more accurate and the model prediction was less accurate than the average human and model guess. There are 1802 and 1721 guesses in the $\ghu\bmo$ and $\bhu\gmo$ settings, respectively. Using these experiments, we aim to understand the effect of explanations on human use and perception of model predictions.

\subsection{How do model predictions and explanation quality affect model-in-the-loop accuracy?}

\paragraph{Participants in the Empathetic, Show Top-3 Range, and Explain-strong treatments performed best and outperformed humans without AI.} 

\begin{figure}[!hbt]
    \centering
    
    \begin{subfigure}{1.0\textwidth}
      \centering
      \includegraphics[width=\textwidth, cfbox=lightgray 0.1pt 0.1pt]{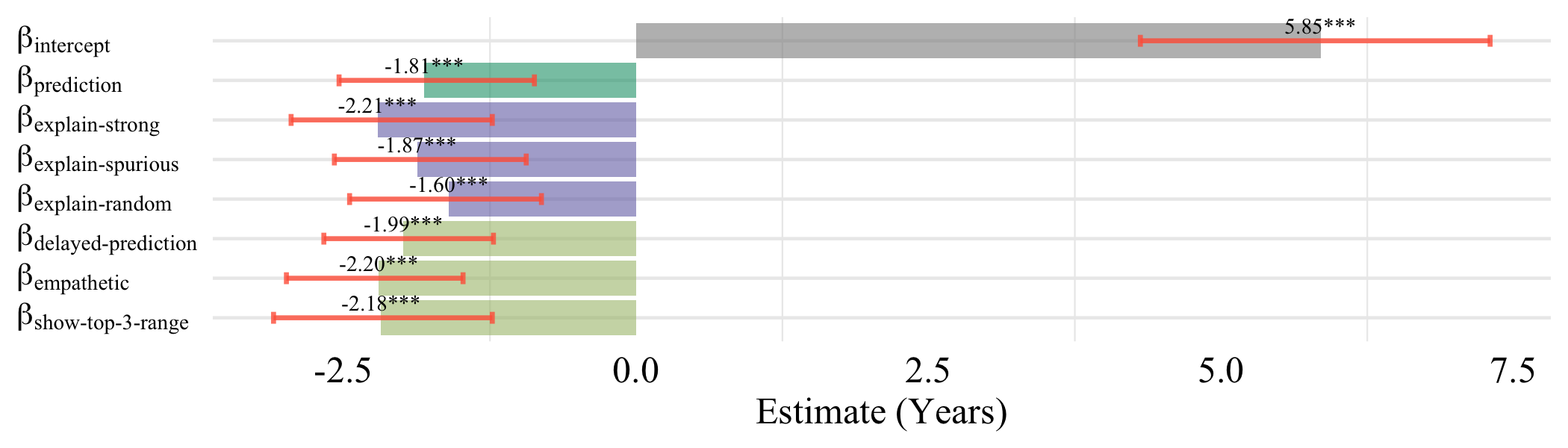}
      \caption{Overall}
      \label{fig:reg-err-overall}
    \end{subfigure}
    \newline
    \begin{subfigure}{.48\textwidth}
      \centering
      \includegraphics[width=\textwidth, cfbox=lightgray 0.1pt 0.1pt]{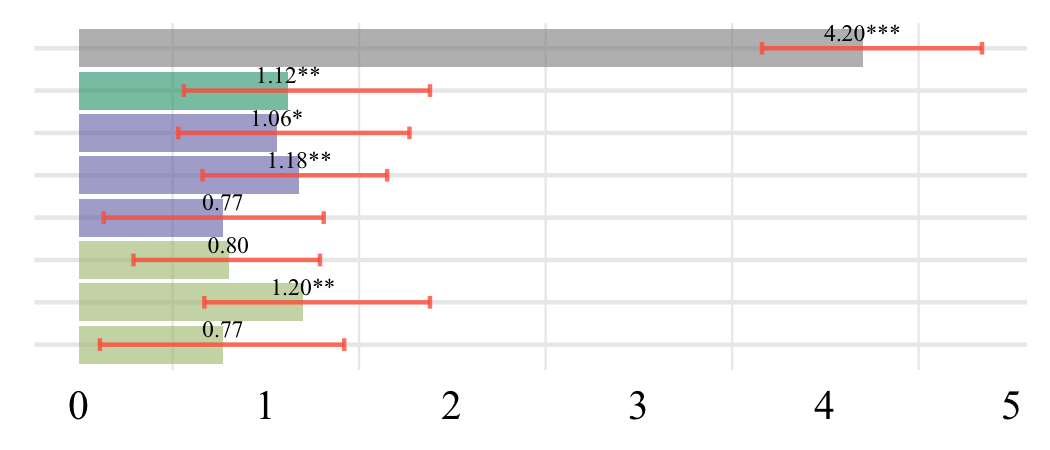}
      \caption{$\ghuman\bmodel$}
      \label{fig:reg-err-hgood-mbad}
    \end{subfigure}
    \hspace{0.02\textwidth}
    \begin{subfigure}{.48\textwidth}
      \centering
      \includegraphics[width=\textwidth, cfbox=lightgray 0.1pt 0.1pt]{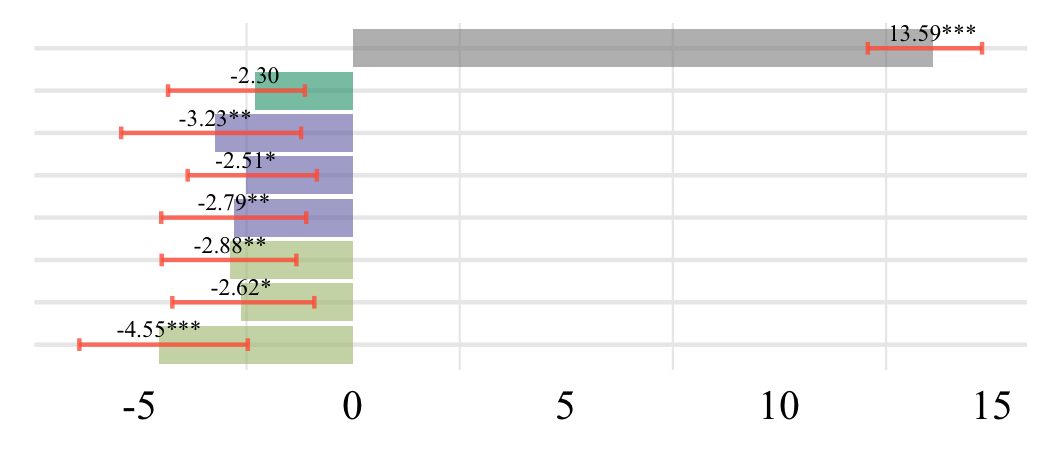}  
      \caption{$\bhuman\gmodel$}
      \label{fig:reg-err-hbad-mgood}
    \end{subfigure}
    \newline
    
    \caption{Estimates for $y_{error}$ regression. \textbf{Intercept represents control treatment (human alone); other estimates are relative to the intercept. Lower values are better, indicating reduced error}. Bootstrapped 95\% confidence intervals are shown. Starred items are statistically significant, calculated with Bonferroni correction: *p<0.05, **p<0.01, ***p<0.001. Additional treatments and settings are in Appendix Section \ref{sec:supp-results}. While showing the model prediction increased accuracy, the explanation-based treatments did not differ significantly from showing the prediction alone. Pairwise differences between the top treatments were also not statistically significant, in all three settings shown.}
    \label{fig:reg-err}
\end{figure}

Participants shown the model prediction are more accurate by 2 years on average overall, as seen in Figure \ref{fig:reg-err}. The predictions generally help whenever the human is inaccurate ($\bhu$$\gmo$, $\bhu$$\bmo$), but can hurt when the human is accurate and the model is inaccurate ($\ghu$$\bmo$).

The top-performing treatments have similar effects overall. However, their effects vary in different settings. For example, Show Top-3 Range is potentially more helpful in the $\bhu\gmo$ setting, with a 4.55 improvement in accuracy, and also the only top treatment to not have a statistically significant harmful effect in the $\ghu\bmo$ setting. However, we note that Tukey HSD testing indicates that the pairwise differences between all of these top treatments is not statistically significant.

The results are also a reminder of the importance of the \textit{design} of the human-AI system, with the Empathetic and Show Top-3 Range treatments equally or more effective as our explanation-based treatments.
Understanding these approaches may also help design better explanation-based approaches, as there may be underlying mechanisms that affect both. For example, the Empathetic and Explain-strong treatments both increase trust, and it could be that the former does so through an emotional approach while the latter does so through more logical means.

Improved accuracy is not attributable to the amount of time spent making guesses:
participants in the Control took 5.7 seconds per image, compared to Empathetic (6.4), Explain-strong (5.9), Show Top-3 Range (5.4), and Prediction (5.2).

\paragraph{The addition of explanations did not improve accuracy.} As seen in Figure \ref{fig:reg-err-overall}, the Explain-strong treatment has a similar effect size to the Prediction treatment. This, along with additional explanation-without-prediction treatments detailed in Appendix Sections \ref{sec:supp-treatments} and \ref{sec:supp-results}, which resulted in small and non statistically significant effects, indicate the limited utility of these explanations for such a visual classification task. Survey responses indicate that participants did indeed examine the highlighted areas and focus on important features such as location-specific wrinkles, but could not extract information that would boost their performance.

It is possible that our results would change if users were extensively trained to interpret and use the saliency maps, instead of only being presented with our short guide.
We do note, however, that prior work on training users to interpret saliency map explanations in a different task did not increase performance when model predictions were also shown \cite{lai2019human, lai2020chicago}. 
We believe nevertheless that one broader takeaway remains the same --- designers of human-AI systems should question the utility of pixel-level saliency explanations when designing ML-powered tools.

\paragraph{The \textit{quality} of explanations had little effect on accuracy.} Though directionally what one might expect overall (explain-strong < explain-spurious < explain-random), the differences are small and not statistically significant in any setting. This is likely related to how explanation quality had little impact on the \textit{trust} participants placed in the model predictions, discussed in the following section.
\subsection{How does explanation quality affect human trust and understanding?}

\begin{figure}[!hbt]
    \centering
    
    \begin{subfigure}{0.68\textwidth}
      \centering
      \includegraphics[width=\textwidth, cfbox=lightgray 0.1pt 0.1pt]{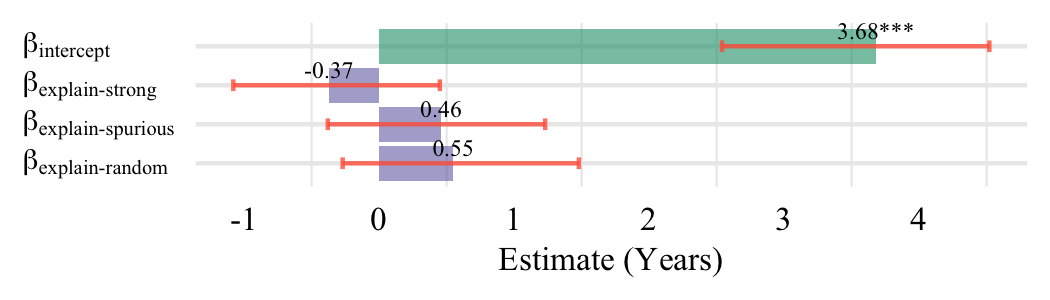}
      \caption{Overall}
      \label{fig:reg-trust-overall}
    \end{subfigure}
    \newline
    \begin{subfigure}{.34\textwidth}
      \centering
      \includegraphics[width=\textwidth, cfbox=lightgray 0.1pt 0.1pt]{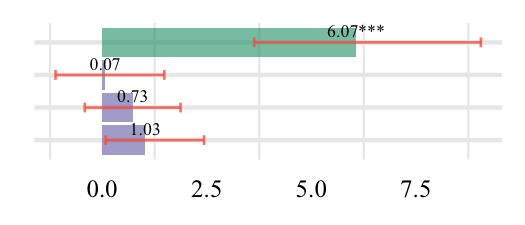}
      \caption{$\ghuman\bmodel$}
      \label{fig:reg-trust-hgood-mbad}
    \end{subfigure}
    \hspace{0.0\textwidth}
    \begin{subfigure}{.34\textwidth}
      \centering
      \includegraphics[width=\textwidth, cfbox=lightgray 0.1pt 0.1pt]{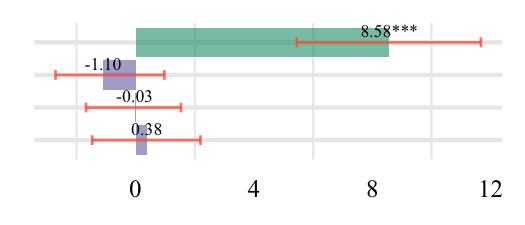}  
      \caption{$\bhuman\gmodel$}
      \label{fig:reg-trust-hbad-mgood}
    \end{subfigure}
    \newline
    
    \caption{Estimates for $y_{trust}$ regression. \textbf{Intercept represents the Prediction treatment; other estimates are relative to the intercept. Smaller values indicate more trust}, i.e. smaller difference between model prediction and human guess. Bootstrapped 95\% confidence intervals are shown. Starred items are statistically significant, calculated with Bonferroni correction: *p<0.05, **p<0.01, ***p<0.001. Additional stats are in Appendix Section \ref{sec:supp-results}. Pairwise differences between explanation-based treatments were not significant, and faulty explanations did not significnatly decrease trust.}
    \label{fig:reg-trust}
\end{figure}

\paragraph{Faulty explanations did not significantly decrease trust in model predictions.} We use the same regression model as in Equation 1 but with ``trust'', the absolute value of the difference between the model's prediction and the user's guess, as the outcome variable. Smaller differences would indicate that explanations can increase the degree to which humans believe model predictions are accurate.
The results are in Figure \ref{fig:reg-trust}.
Strong explanations could \textit{increase} trust up to 1.08 years (CI lower bound) relative to the Prediction treatment (no explanations), while the random explanations could \textit{decrease} trust up to 1.48 years (CI upper bound). These effect sizes could be sizable (30-40\% relative to the intercept mean), but no treatment was statistically significant.
Moreover, the difference between the spurious and random saliency maps is small, and none of the pairwise differences are statistically significant.
These findings hold even when the model prediction is inaccurate ($\bmodel$), indicating
that users are not learning to trust the model based on its predictions alone.

These findings, coupled with the large differences in accuracy between the $\ghu\bmo$ and $\bhu\gmo$ settings and the decrease in accuracy in the $\ghu\bmo$ setting (Figure \ref{fig:reg-err}), 
raise the question to what degree humans can identify and ignore erroneous model predictions.
For example, a model that is accurate 98\% of the time but makes large errors in the remaining 2\% can be very dangerous in a high-stakes scenario if humans default to trusting all model predictions. Importantly, these subsets are typically unknown on test sets. Two possible approaches to alleviate this problem are (a) surfacing uncertainty measures as touched upon in the Show Top-3 Range treatment, and (b) training users by showing common model errors on a validation set. We also briefly expand upon the complementarity of human guesses and model decisions in Appendix Section \ref{sec:supp-combine}. 

\begin{wrapfigure}{R}{0.4\textwidth}
    \centering
    \includegraphics[width=0.4\textwidth]{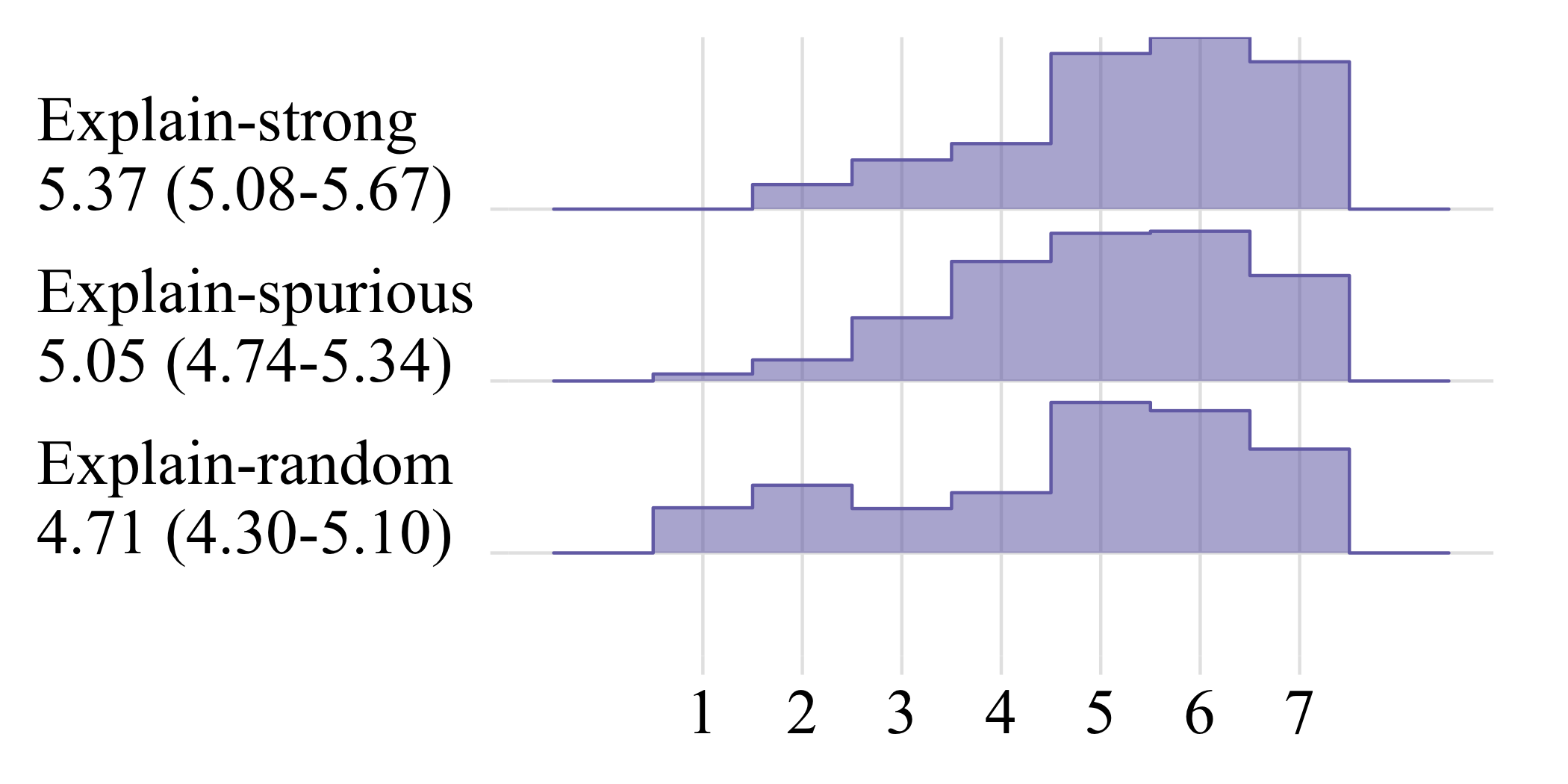}
    \caption{Distribution, mean, and 95\% confidence intervals on the intelligiblity of explanations. Response on a 1-7 Likert scale. Participants rated explanations similarly, regardless of the quality of the explanations.}
    \label{fig:att1}
\end{wrapfigure}

\paragraph{Most participants claimed that explanations appeared reasonable, \textit{even when they were obviously not focused on faces.}} 
Responses out of 7 to the post-survey question, ``How easy to understand were the AI's explanations? Did the explanations seem reasonable?'', are shown in Figure \ref{fig:att1}. 
Despite the clear flaws in the saliency maps shown in Figure \ref{fig:img-atts}, there were only small differences between the treatments. Participants shown a strong explanation rated the explanations 5.37 / 7, versus 5.05 and 4.71 for the spurious and random explanations. 
In the next section, we provide qualitative examples to shed some light into these counterintuitive findings.

\subsection{How did humans incorporate model predictions and explanations into their decision making process?}

Responses to the post-survey question ``How did you use the model's prediction in your decision-making process?'' provide some clues to the workings of our human-AI system. Participants in the Explain-random treatment did highlight the randomness of the saliency maps, with one saying ``I took it slightly into consideration but didn't weight it heavily because it looked like the model was picking inaccurate locations...''. However, many others focused simply on the accuracy of the prediction, with one stating ``Well it did a poor job recognizing face or features but the ages sound mostly correct so i sort of went with it''. We again note, however, that faulty explanations did not significantly decrease trust even in the presence of \textit{inaccurate} model predictions (Figure \ref{fig:reg-err-hgood-mbad}). Participants in the Explain-spurious treatment were similar, sometimes noting that the explanations were ``totally out of whack, like on the wall'', but with only a few, explicit statements of these explanations mediating judgment, such as ``If it was close to an area that might actually matter (neck, under eyes, edges of mouth etc) I took it into consideration, but if not, I dismissed it.''

We also examined responses for the top 75 guessers in terms of mean error, (collective MAE of 5.19). Answers to how the model prediction was used were bucketed into 6 categories: (1) 10.3\% ignored it, (2) 17.9\% considered it, (3)  5.1\% used it if they were unsure, (4) 28.2\% used the model prediction as a starting point, (5) 21.8\% had their own initial guess and adjusted based on the model prediction, (6) 9.0\% used the model prediction if it seemed reasonable; otherwise gave their own guess. 

\section{Related Work}
\label{sec:related}




\paragraph{Interpretable machine learning}

There has been a wide range of methods, including instance-level explanations \cite{ribeiro2016should} vs. global explanations based on feature representations across the dataset \cite{bau2017network}, glass-box methods with access to model gradients \cite{sundararajan2017axiomatic,smilkov2017smoothgrad} vs. black box methods \cite{ribeiro2016should,petsiuk2018rise}, and input feature attributions \cite{ribeiro2016should,sundararajan2017axiomatic,smilkov2017smoothgrad,petsiuk2018rise,bau2017network} vs. natural language explanations \cite{hendricks2016generating,lei2016rationalizing} vs. counterfactuals \cite{wachter2017counterfactual}. Input feature attributions has been a common approach, which we use for our experiment.

Evaluating explanations has included human assessments of correctness, attempts to derive insights from explanations, and seeing if explanations help humans predict model behavior \cite{sundararajan2017axiomatic, ribeiro2016should, chandrasekaran2017takes, narayanan2018humans}. Recent work has also suggested that popular interpretability methods exhibit undesirable properties, such as being unreliably and unreasonably sensitive to minor changes in the input \cite{jain2019attention,kindermans2019reliability,feng2019can}.


\paragraph{Model-in-the-Loop experiments}

Previous experiments have mixed results on the effect of explanations. In a deceptive review detection task, explanations alone helped human performance slightly. However, the best setting was simply showing the prediction alone while informing users of the model's accuracy \cite{lai2019human}. Other work has shown feature-wise importances for low-dimensional inputs to both slightly increase \cite{green2019principles} and decrease accuracy \cite{zhang2020effect}. Explanations have also been found to increase trust and understanding \cite{lai2019human, lim2009and}, but can hurt if they too overtly convey the model's limitations \cite{yeomans2019making}. Model complexity was also examined in \cite{poursabzi2018manipulating}, which found that less complex, transparent, linear models did not help users in an apartment price prediction task. In fact, transparency hindered people from detecting model errors, echoing other work on the risk of cognitive overload \cite{abdulcogam, lage2019human}.

We expand upon these prior works, as their limitations include (a) linear, relatively simple, or non-neural models \cite{poursabzi2018manipulating, feng2019can, binns2018s, lage2019human, lai2019human}, (b) small input feature spaces, making explanations simpler \cite{green2019principles, poursabzi2018manipulating, zhang2020effect}, (c) imbalance of task performance between humans and AI, e.g. human performance is 51\% on a binary classification task (near random), vs. 87\% for the model in \cite{lai2019human, lai2020chicago}, (d) no investigation into using explanations for certification (identifying faulty or biased models) \cite{ binns2018s, feng2019can, green2019principles,  lai2020chicago, lai2019human,poursabzi2018manipulating, tan2018investigating, zhang2020effect}.

\paragraph{Certification with explanations} In an ad recommendation setting, explanations allowed users to detect that models were highly limited \cite{eslami2018communicating}. 
The authors of \cite{ribeiro2016should} found users able to identify the better of two models using their interpretable method with 90\% accuracy. \cite{adebayo2018sanity} introduces ``model parameter randomization'' and ``data randomization'' tests to analyze whether explanations are specific to the model and input. However, there have not been extensive human studies of certification. 





\paragraph{Design of the human-AI system}
Varying the stated accuracy of the model can greatly increase trust in the model, though this trust will decrease if the observed accuracy is low \cite{yin2019understanding}. There are also potential benefits for displaying uncertainty estimates for each prediction \cite{van2019communicating}, though \cite{zhang2020effect} found no improvement when showing confidence scores in textual form. 
``Design''
covers an even broader category of elements, such as incentives to use a ML-driven tool, level of human agency and interactivity, and the ``unremarkability'' of the system \cite{yang2019unremarkable}. These can often determine success in real-life settings \cite{cai2019human, sendak2020human, cai2019hello, beede2020human}, but are out of the scope of our study.


\section{Conclusion}
Ideally, contextualizing model predictions with explanations would help improve people's decision-making process in model-in-the-loop settings. Randomized control trials on this age prediction task, however,
found that additionally showing explanations with model predictions did not have a significant effect on accuracy, trust, or understanding. Moreover, the quality of the explanations were unimportant, and even faulty explanations did not significnatly decrease human trust in the model.

Existing interpretable ML methods have largely been used as a debugging tool for researchers and industry engineers, rather than a mechanism for communicating with end users \cite{bhatt2020explainable}. Our findings are a reminder of this point and suggest input feature attributions, while helpful for machine learning researchers, may be less useful in downstream decision-making problems. 
Other interpretability methods, such as counterfactuals or natural language explanations, may be more effective but should be properly motivated and evaluated by their downstream use if model-in-the-loop usage is a potential goal.
This echoes broader trends in human-computer interaction \cite{harrison2011making, harrison2007three} and grounded ML research, which argue for greater situating of ML models and methods in real-world scenarios.

\section*{Broader Impact}
We chose our image classification task precisely because there are analogus models used in high-stakes situations. There are risks to examining such a problem, as these findings could also be used to improve ethically-dubious systems involving prediction of demographic attributes such as facial recognition for surveillance systems or criminial identification. Complementary human and model biases may be also be amplified in a human-AI system, further harming already disproportionately marginalized communities.

However, we believe the machine learning community and designers of ML-powered tools may immediately benefit from the findings of our study, as it motivates more useful explanations and ways to design human-AI systems. Ultimately, we hope this will improve the legibility of automated deicision systems for non-technical users and improve outcomes for those affected by ML-powered tools.


\section*{Acknowledgements}
Thank you to Nabeel Gillani, Martin Saveski, Doug Beeferman, Sneha Priscilla Makini, and Nazmus Saquib for helpful discussion about the design and analysis of the RCT, as well as Jesse Mu for feedback on the paper.


\bibliographystyle{plain}
\bibliography{neurips_2020}

\iftoggle{arxiv}{
        
    \newpage
    \appendix
    \setcounter{section}{0}
    
    \setcounter{table}{0}
    \renewcommand{\thetable}{S\arabic{table}}%
    \setcounter{figure}{0}
    \renewcommand{\thefigure}{S\arabic{figure}}%
    

%
%

\section{Experimental Setup}

\subsection{Datasets, Models, and Training Details}
\label{sec:supp-models}
The IMDB-WIKI dataset used to pretrain our strong model consists of 524,230 images and age labels, and was partitioned into 90\% train, 5\% validation, and 5\% test splits. The train, validation, and test splits for the APPA-REAL dataset were included in the original dataset\footnote{Download link: \url{http://chalearnlap.cvc.uab.es/dataset/26/data/45/description/}}. We also trained a ``weak'' model for use in one additional treatment, which was not pre-trained on the IMDB-WIKI dataset and not trained until convergence on the APPA-REAL dataset. Saliency maps from the weak model tended to focus on similar features, but were often more diffuse, with fewer red spots. The strong and weak models used pretrained ImageNet weights, while the spurious model was trained from scratch in order to be more tuned to the spurious correlations. The model MAEs are listed in Table \ref{tab:age-models-stats}.

\begin{table}[h]
    \caption{Age prediction model performance in terms of mean absolute error (MAE). The MAE for the Spurious model is on the modified APPA-REAL dataset and has slightly lower MAE than the Strong model precisely because it is tuned into additional spurious correlations.}
    \label{tab:age-models-stats}
    \vspace{0.5em}

  \centering
  \begin{tabular}{lcc}
    \toprule
    Model & Valid MAE & Test MAE \\
    \midrule
    Strong & 4.62 & 5.24 \\
    Weak & 5.84 & 6.86 \\
    Spurious & 3.33 & 4.58 \\
    \bottomrule
  \end{tabular}
  
\end{table}

Each model was trained on one GeForce GTX 1080 Ti using Pytorch. We used the Adam optimizer with learning rate 0.001 (after sweeping over [0.01, 0.005, 0.001, 0.0005]). Images were scaled to 224 $\times$ 224. We performed data augmentation during training by applying additive Gaussian noise 12.5\% of the time, Gaussian blur 12.5\% of the time, 20 degree rotations, 5\% scaling and translation operations, horizontal flipping, and hue and saturation adjustments.

\subsection{Treatment Arms}
\label{sec:supp-treatments}

The avatar used for the \textit{Empathetic} treatment is shown in Figure \ref{fig:pat}. The icon was made by Freepik from \url{www.flaticon.com} and can be found at \url{https://www.flaticon.com/free-icon/robot_1587565}.
\begin{figure}[!hbt]
    \centering
    \includegraphics[width=0.1\textwidth]{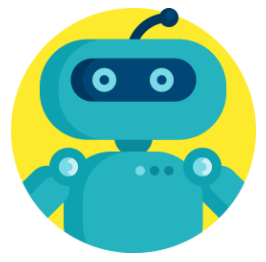}
    \caption{Pat the AI.}
    \label{fig:pat}
\end{figure}

\paragraph{Additional treatments}
We also tested the effect of explanations alone, hypothesizing that they may have a small, but slight effect.
The \textit{Explain-strong, No Pred} and \textit{Explain-weak, No Pred} treatments show the saliency maps from the strong and weak model, respectively, \textit{without the prediction}.

We also tested a global, model-level explanation in the \textit{Explain-guide, No Pred} treatment. Before the task begins, users are shown a grid of saliency maps and told that important regions are: ``(1) the areas around the eyes, and (2) lines anywhere on the face. The next two most important regions are around the mouth and nose.'' The researchers manually went through 200 saliency maps and tallied regions in red in order to determine these features. Users are reminded of these guidelines at every image. This approach is similar in spirit to \cite{lai2020chicago}.


For the faulty saliency maps, we additionally tested not stating the model's performance in the \textit{No Acc} treatments.
We hypothesized that allowing users to come to their own conclusion about the model's ability would result in the faulty explanations having a larger effect.

\begin{table}[h]

  \caption{Full list of treatment arms. All model predictions are from the same ``strong'' model.}
  \label{tab:supp-treatments}
  \vspace{0.5em}
  
  \centering
  \begin{tabular}{lll}
    \toprule
    & \bf Treatment Arm & \bf Shorthand Description\\
    \toprule
    & Control & Ask for age without help of model \\
    \midrule
    & Prediction & User is shown model's prediction \\
    
    \midrule
    \multirow{3}{*}{\rotatebox[origin=c]{90}{\emph{Design}}}
    & Delayed Prediction & User guesses before and after seeing model's prediction \\
    & Empathetic & Model is described as a fallible ``Pat the AI''  \\
    & Show Top-3 Range & Prediction shown as range of top-3 values, e.g. 28-32 \\

    \midrule
    \multirow{7}{*}{\rotatebox[origin=c]{90}{\emph{Explanations}}}
    & Explain-strong & Show strong model's saliency map \\
    & Explain-spurious & Show spurious model's saliency map \\
    \vspace{0.5em}
    & Explain-random & Show random saliency map \\
    & Explain-strong, No Pred & Show strong model's saliency map, hide prediction \\
    & Explain-weak, No Pred & Show weak model's saliency map, hide prediction \\
    \vspace{0.5em}
    & Explain-guide, No Pred & Show summary of feature importances, hide prediction \\
    
    & Explain-spurious, No Acc & Show spurious model's saliency map, hide model's accuracy  \\
    & Explain-random, No Acc & Show random saliency map, hide model's accuracy \\
    \bottomrule
  \end{tabular}
  
\end{table}

\subsection{Participant Demographics}
40.3\% of the participants were female, with a mean age of 36.5 (standard deviation 10.8).

%
%

\section{Results}
\label{sec:supp-results}

The full regression results for MAE, accuracy, and trust are shown in Tables \ref{tab:supp-mae}, \ref{tab:supp-err-reg}, and \ref{tab:supp-trust-reg}. We briefly cover findings from the additional treatments as follows:

\begin{itemize}
    \item The instance-level explanation-only treatments (\textit{Explain-strong, No Pred and Explain-weak, No Pred}) had small, non-statistically significant effects on accuracy.
    \item The model-level explanation-only treatment, \textit{Explain-guide, No Pred}, was helpful overall (1.2 years increase in accuracy, compared to 1.8 years increase in the \textit{Prediction} treatment). It was also as helpful as the top performing treatments when both humans and models were inaccurate, i.e. in the $\bhuman\bmodel$ setting.
    \item Contrary to our hypothesis, hiding the model performance did not significantly increase or decrease the effect of the faulty explanations. Directionally, however, it appeared to \textit{increase trust} in the $\bhuman\bmodel$ setting. It may be that the statement on model performance actually emphasized the fallability of the model.
\end{itemize}



\begin{table}[!htbp]

    \caption{Mean absolute error of guesses for all treatments. Bootstrapped 95\% confidence intervals in parentheses. 
    }
   \label{tab:supp-mae}
   \vspace{0.5em}
   
      \centering
      \begin{tabular}{lc}
        \toprule
        Treatment Arm & MAE \\
        \midrule
        Control (Human Alone) & 10.0 (9.4 - 10.5) \\
        Model Alone & 8.5 (8.3 - 8.7) \\
        Prediction & 8.4 (7.8 - 9.0) \\
        \midrule
        Explain-strong & 8.0 (7.5 - 8.5) \\
        Explain-spurious & 8.5 (8.0 - 9.1) \\
        Explain-random & 8.7 (8.1 - 9.2) \\
        \midrule
        Delayed Prediction & 8.5 (8.0 - 9.0) \\
        Empathetic & 8.0 (7.6 - 8.5) \\
        Show Top-3 Range & 8.0 (7.4 - 8.5) \\
        \midrule
        Explain-strong, No Pred & 9.7 (9.2 - 10.3) \\
        Explain-weak, No Pred & 10.2 (9.5 - 10.9) \\
        Explain-guide, No Pred & 9.4 (8.9 - 10.0) \\
        \midrule
        Explain-spurious, No Acc & 8.2 (7.6 - 8.8) \\
        Explain-random, No Acc & 8.0 (7.6 - 8.5) \\
        \bottomrule
      \end{tabular}
\end{table}

\begin{table}
      \caption{Estimates for $y_{error}$ regression. \textbf{Intercept represents control treatment (human alone); other estimates are relative to the intercept. Lower values are better, indicating reduced error}. Bootstrapped 95\% confidence intervals are shown in parentheses; starred items are statistically significant, calculated with Bonferroni correction: *p<0.05, **p<0.01, ***p<0.001.}
      \label{tab:supp-err-reg}
      \vspace{0.5em}
      
      \begin{subtable}{1.0\textwidth}
      
      \centering
      \begin{tabular}{Ll}
      
        \toprule
         & \multicolumn{1}{l}{Overall} \\
        \midrule 
        
        \beta_{\textrm{intercept}} & \multicolumn{1}{l}{5.9 (4.6,7.2)***}\\
        \midrule
        \beta_{\textrm{prediction}} & \multicolumn{1}{l}{-1.8 (-2.6,-1.1)***}\\
        \midrule
        \beta_{\textrm{delay-model-pred}} & \multicolumn{1}{l}{-2.0 (-2.8,-1.2)***}\\
        \beta_{\textrm{empathetic}} & \multicolumn{1}{l}{-2.2 (-3.0,-1.2)***}\\
        \beta_{\textrm{show-top-3-range}} & \multicolumn{1}{l}{-2.2 (-2.9,-1.3)***}\\
        \midrule
        \beta_{\textrm{explain-strong}} & \multicolumn{1}{l}{-2.2 (-3.2,-1.3)***}\\
        \beta_{\textrm{explain-spurious}} & \multicolumn{1}{l}{-1.9 (-2.7,-1.2)***}\\
        \beta_{\textrm{explain-random}} & \multicolumn{1}{l}{-1.6 (-2.5,-0.9)***}\\
        \midrule
        \beta_{\textrm{explain-strong\_no-pred}} & \multicolumn{1}{l}{-0.5 (-1.5,0.3)}\\
        \beta_{\textrm{explain-weak\_no-pred}} & \multicolumn{1}{l}{0.2 (-0.7,0.9)}\\
        \beta_{\textrm{explain-guide\_no-pred}} & \multicolumn{1}{l}{-1.2 (-2.0,-0.4)*}\\
        \midrule
        \beta_{\textrm{explain-spurious\_no-acc}} & \multicolumn{1}{l}{-1.9 (-2.7,-1.0)***}\\
        \beta_{\textrm{explain-random\_no-acc}} & \multicolumn{1}{l}{-2.0 (-2.6,-1.2)***}\\
        \midrule
        \beta_{\textrm{image-age}} & \multicolumn{1}{l}{0.1 (0.0,0.1)***}\\
        
        \bottomrule
      \end{tabular}
      \caption{Overall}
      \end{subtable}
      
      \begin{subtable}{1.0\textwidth}
      \small
      \centering
      \begin{tabular}{Lllll}
        \toprule
         & \multicolumn{1}{l}{$\ghuman\gmodel$} 
         & \multicolumn{1}{l}{$\ghuman\bmodel$}
         & \multicolumn{1}{l}{$\bhuman\gmodel$} 
         & \multicolumn{1}{l}{$\bhuman\bmodel$} \\
        \midrule 
        N & 4940 & 1802 & 1721 & 2812 \\
        \toprule
         
        \beta_{\textrm{intercept}} & \multicolumn{1}{l}{3.3 (2.9,3.7)***} & \multicolumn{1}{l}{4.2 (3.6,4.8)***} & \multicolumn{1}{l}{13.6 (12.1,14.8)***} & \multicolumn{1}{l}{12.7 (10.3,15.9)***}\\
        \midrule
        \beta_{\textrm{prediction}} & \multicolumn{1}{l}{-0.3 (-0.6,0.0)} & \multicolumn{1}{l}{1.1 (0.5,1.8)**} & \multicolumn{1}{l}{-2.3 (-4.3,-1.1).} & \multicolumn{1}{l}{-2.0 (-3.4,-0.7)*}\\
        \midrule
        \beta_{\textrm{delay-model-pred}} & \multicolumn{1}{l}{-0.1 (-0.4,0.3)} & \multicolumn{1}{l}{0.8 (0.3,1.3).} & \multicolumn{1}{l}{-2.9 (-4.5,-1.3)**} & \multicolumn{1}{l}{-2.3 (-3.5,-1.0)**}\\
        \beta_{\textrm{empathetic}} & \multicolumn{1}{l}{-0.4 (-0.7,-0.0)} & \multicolumn{1}{l}{1.2 (0.6,2.0)**} & \multicolumn{1}{l}{-2.6 (-4.2,-0.9)*} & \multicolumn{1}{l}{-2.7 (-3.7,-1.6)***}\\
        \beta_{\textrm{show-top-3-range}} & \multicolumn{1}{l}{-0.4 (-0.8,-0.0)} & \multicolumn{1}{l}{0.8 (0.3,1.4)} & \multicolumn{1}{l}{-4.5 (-6.4,-2.5)***} & \multicolumn{1}{l}{-1.7 (-2.8,-0.8)*}\\
        \midrule
        \beta_{\textrm{explain-strong}} & \multicolumn{1}{l}{-0.2 (-0.6,0.1)} & \multicolumn{1}{l}{1.1 (0.5,1.6)*} & \multicolumn{1}{l}{-3.2 (-5.4,-1.2)**} & \multicolumn{1}{l}{-2.4 (-3.4,-1.2)**}\\
        \beta_{\textrm{explain-spurious}} & \multicolumn{1}{l}{-0.2 (-0.5,0.1)} & \multicolumn{1}{l}{1.2 (0.7,1.7)**} & \multicolumn{1}{l}{-2.5 (-3.9,-0.9)*} & \multicolumn{1}{l}{-1.5 (-2.8,-0.3)}\\
        \beta_{\textrm{explain-random}} & \multicolumn{1}{l}{-0.2 (-0.5,0.2)} & \multicolumn{1}{l}{0.8 (0.2,1.3)} & \multicolumn{1}{l}{-2.8 (-4.5,-1.1)**} & \multicolumn{1}{l}{-1.7 (-3.0,-0.5).}\\
        \midrule
        \beta_{\textrm{explain-strong\_no-pred}} & \multicolumn{1}{l}{0.4 (0.0,0.9)} & \multicolumn{1}{l}{0.2 (-0.3,0.8)} & \multicolumn{1}{l}{-1.1 (-2.8,0.4)} & \multicolumn{1}{l}{-1.4 (-2.7,0.0)}\\
        \beta_{\textrm{explain-weak\_no-pred}} & \multicolumn{1}{l}{0.1 (-0.3,0.7)} & \multicolumn{1}{l}{-0.1 (-0.8,0.8)} & \multicolumn{1}{l}{1.0 (-0.9,2.4)} & \multicolumn{1}{l}{-0.6 (-1.8,0.9)}\\
        \beta_{\textrm{explain-guide\_no-pred}} & \multicolumn{1}{l}{0.0 (-0.2,0.4)} & \multicolumn{1}{l}{-0.2 (-0.8,0.3)} & \multicolumn{1}{l}{-1.1 (-2.6,0.8)} & \multicolumn{1}{l}{-2.4 (-3.9,-1.1)**}\\
        \midrule
        \beta_{\textrm{explain-spurious\_no-acc}} & \multicolumn{1}{l}{-0.3 (-0.7,0.1)} & \multicolumn{1}{l}{1.1 (0.5,1.9)**} & \multicolumn{1}{l}{-2.3 (-3.7,-0.8)} & \multicolumn{1}{l}{-1.2 (-2.4,0.1)}\\
        \beta_{\textrm{explain-random\_no-acc}} & \multicolumn{1}{l}{-0.2 (-0.5,0.3)} & \multicolumn{1}{l}{1.1 (0.6,1.8)**} & \multicolumn{1}{l}{-3.0 (-4.7,-1.6)**} & \multicolumn{1}{l}{-2.2 (-3.2,-1.1)**}\\
        \midrule
        \beta_{\textrm{image-age}} & \multicolumn{1}{l}{0.0 (0.0,0.0)*} & \multicolumn{1}{l}{-0.0 (-0.0,0.0)} & \multicolumn{1}{l}{0.0 (0.0,0.1)} & \multicolumn{1}{l}{0.1 (0.0,0.1)*}\\
        
        \bottomrule
      \end{tabular}
      \caption{Splits}
      \end{subtable}
\end{table}


%
%


\begin{table}
      \caption{Estimates for $y_{trust}$ regression. \textbf{Intercept represents Prediction treatment; other estimates are relative to the intercept. Lower values indicate greater trust}. Bootstrapped 95\% confidence intervals are shown in parentheses; starred items are statistically significant, calculated with Bonferroni correction: *p<0.05, **p<0.01, ***p<0.001.}
      \label{tab:supp-trust-reg}
      \vspace{0.5em}
      
      \begin{subtable}{1.0\textwidth}
      \centering
      \begin{tabular}{Ll}
      
        \toprule
         & \multicolumn{1}{l}{Overall} \\
        \midrule 
        
        \beta_{\textrm{intercept}} & \multicolumn{1}{l}{3.7 (2.5,4.5)***}\\
        \midrule
        \beta_{\textrm{explain-strong}} & \multicolumn{1}{l}{-0.4 (-1.1,0.4)}\\
        \midrule
        \beta_{\textrm{explain-spurious}} & \multicolumn{1}{l}{0.5 (-0.4,1.2)}\\
        \beta_{\textrm{explain-random}} & \multicolumn{1}{l}{0.5 (-0.3,1.5)}\\
        \midrule
        \beta_{\textrm{explain-spurious\_no-acc}} & \multicolumn{1}{l}{0.3 (-0.5,1.0)}\\
        \beta_{\textrm{explain-random\_no-acc}} & \multicolumn{1}{l}{0.4 (-0.5,1.0)}\\
        \midrule
        \beta_{\textrm{image-age}} & \multicolumn{1}{l}{0.0 (0.0,0.0)*}\\
        
        \bottomrule
      \end{tabular}
      \caption{Overall}
      \end{subtable}
      
      \begin{subtable}{1.0\textwidth}
      \centering
      \begin{tabular}{Lllll}
        \toprule
         & \multicolumn{1}{l}{$\ghuman\gmodel$} 
         & \multicolumn{1}{l}{$\ghuman\bmodel$}
         & \multicolumn{1}{l}{$\bhuman\gmodel$} 
         & \multicolumn{1}{l}{$\bhuman\bmodel$} \\
        \midrule 
        N & 4940 & 1802 & 1721 & 2812 \\
        \toprule

        \beta_{\textrm{explain-strong}} & \multicolumn{1}{l}{-0.1 (-0.5,0.5)} & \multicolumn{1}{l}{0.1 (-1.1,1.5)} & \multicolumn{1}{l}{-1.1 (-2.7,1.0)} & \multicolumn{1}{l}{0.1 (-1.5,1.6)}\\
        \midrule
        \beta_{\textrm{explain-spurious}} & \multicolumn{1}{l}{0.1 (-0.3,0.7)} & \multicolumn{1}{l}{0.7 (-0.4,1.9)} & \multicolumn{1}{l}{-0.0 (-1.7,1.5)} & \multicolumn{1}{l}{1.0 (-0.3,2.1)}\\
        \beta_{\textrm{explain-random}} & \multicolumn{1}{l}{0.0 (-0.4,0.4)} & \multicolumn{1}{l}{1.0 (0.1,2.4)} & \multicolumn{1}{l}{0.4 (-1.5,2.2)} & \multicolumn{1}{l}{1.4 (-0.2,2.6)}\\
        \midrule
        \beta_{\textrm{explain-spurious\_no-acc}} & \multicolumn{1}{l}{0.3 (-0.1,0.7)} & \multicolumn{1}{l}{0.7 (-0.2,1.9)} & \multicolumn{1}{l}{1.2 (-1.2,3.1)} & \multicolumn{1}{l}{-0.2 (-1.8,1.1)}\\
        \beta_{\textrm{explain-random\_no-acc}} & \multicolumn{1}{l}{0.3 (-0.1,0.8)} & \multicolumn{1}{l}{0.7 (-0.3,1.9)} & \multicolumn{1}{l}{-0.3 (-2.0,1.5)} & \multicolumn{1}{l}{0.2 (-1.2,1.2)}\\
        \midrule
        \beta_{\textrm{image-age}} & \multicolumn{1}{l}{0.0 (-0.0,0.0)} & \multicolumn{1}{l}{0.0 (-0.0,0.1)} & \multicolumn{1}{l}{0.0 (-0.0,0.1)} & \multicolumn{1}{l}{0.0 (-0.0,0.1)}\\
        
        \bottomrule
      \end{tabular}
      \caption{Splits}
      \end{subtable}
      
\end{table}

%
%

\clearpage

\section{Additional Analyses}

\subsection{The \textit{economic cost} of Model-in-the-Loop predictions}

We also consider the \textit{economic cost} of a prediction, calculated simply as error $\times$ time. Under this metric, the treatment arms are similar or worse than simply showing the model prediction: Control (56.7), Empathetic (50.6), Explain-strong (47.2), Show Model Pred (43.5), Show Top-3 Range (42.5). We use this simply as an illustrative example, as we believe time and cognitive load are important considerations when designing human-AI systems, especially with possibly complex explanations of high-dimensional inputs.

\subsection{Combining human guesses and model predictions}
\label{sec:supp-combine}

We investigate the possible gain of combining the two predictions in simple hybrid models, whose input features are simply the model's prediction and the human's guess. Such models have been found to outperform either human or machine alone \cite{wang2016deep, phillips2018face}, though this may not always be the case \cite{tan2018investigating}. We perform cross-fold validation and hyperparameter search, with the test MAE results for the top-performing treatments shown in Table \ref{tab:hybrid-models}. 

Follow-up questions to our work include: (a) how the differences between treatments can be related to the complementarity of human and model predictions, and (b) how to best design human-AI systems where the AI complements existing human capabilities.
One experiment could be to derive coarse strategies from these hybrid models, such as important decision tree rules, and test whether these strategies could help further improve accuracy (i.e. ``Trust the model if you think the person is above 60, and the model's prediction is significantly greater than yours.'')

\begin{table}[!hbtp]
  \caption{MAE on test split of hybrid models that combine human guesses and model predictions}
  \label{tab:hybrid-models}
  \vspace{0.5em}
  
  \small
  \centering
  \begin{tabular}{lS[round-precision=\precision]S[round-precision=\precision]S[round-precision=\precision]S[round-precision=\precision]}
    \toprule
    & \multicolumn{1}{c}{Prediction} & \multicolumn{1}{c}{Explain-strong} & \multicolumn{1}{c}{Empathetic} & \multicolumn{1}{c}{Show Top-3 Range}\\
    \midrule 
    Human guess (with AI assistance) & 8.3910 & 7.9720 & 7.9894 & 7.9780  \\
    Model prediction & 8.4110 & 8.392 & 8.3181 & 8.5152 \\
    \midrule
    Logistic Regression & 8.0875 & 8.3333 & 7.4302 & 9.3780 \\
    Decision Tree & 7.1375 &  6.2933 & 8.3721 & 5.9878 \\
    \bottomrule
  \end{tabular}
\end{table}



}

\end{document}